\documentclass[review,a4paper,fleqn]{cas-sc}
\usepackage[numbers,sort&compress]{natbib}
\usepackage{fancyhdr}
\usepackage{amsmath}
\usepackage{algorithm}
\usepackage{algpseudocode}
\usepackage{amsfonts}
\usepackage{booktabs}
\usepackage{tabularx}
\usepackage{graphicx}
\usepackage{chngcntr}
\usepackage{hyperref}
\counterwithout{table}{section}
\usepackage{nomencl}
\makenomenclature
\usepackage{etoolbox}
 
\renewcommand{\nompreamble}{\begin{multicols}{2}}
\renewcommand{\nompostamble}{\end{multicols}}
\usepackage{framed}
\usepackage{multicol}
\usepackage{amsmath,amssymb}
\usepackage{siunitx}
\usepackage{setspace}
\makeatletter

\nomenclature{AI}{Artificial Intelligence}
\nomenclature{DRL}{Deep Reinforcement Learning}
\nomenclature{MDP}{Markov Decision Process}
\nomenclature{DLCF}{Dual-Loop Control Framework}
\nomenclature{SLCF}{Single-Loop Control Framework}
\nomenclature{PIML}{Physics-Informed Machine Learning}
\nomenclature{CRAH}{Computer Room Air Handler}
\nomenclature{CHW}{Chilled Water}
\nomenclature{ICT}{Information and Communication Technology}
\nomenclature{SLA}{Service Level Agreement}
\nomenclature{MAPE}{Mean Absolute Percentage Error}
\nomenclature{O\&M}{Operation and Maintenanc}
\nomenclature{MPC}{Model Predictive Control}
\nomenclature{DNN}{Deep Neural Network}
\nomenclature{$\mathcal{S}$}{State space}
\nomenclature{$\mathcal{A}$}{Action space}
\nomenclature{$M$}{Transition dynamics model}
\nomenclature{$R$}{Reward function}
\nomenclature{$\pi$}{Policy function}
\nomenclature{$\gamma$}{Discount factor}
\nomenclature{$V^\pi(s)$}{State value function under policy $\pi$}
\nomenclature{$Q^\pi(s,a)$}{State-action value function under $\pi$}
\nomenclature{$\rho^\pi$}{State-action distribution under policy $\pi$}

\def\tsc#1{\csdef{#1}{\textsc{\lowercase{#1}}\xspace}}
\tsc{WGM}
\tsc{QE}
\tsc{EP}
\tsc{PMS}
\tsc{BEC}
\tsc{DE}

\begin{document}
\let\WriteBookmarks\relax
\def\floatpagepagefraction{1}
\def\textpagefraction{.001}
\doublespacing

\title [mode = title]{Dual-Loop Control in DCVerse: Advancing Reliable Deployment of AI in Data Centers via Digital Twins}

\author[1]{Qingang Zhang}
\ead{qingang.zhang@ntu.edu.sg}
\credit{Writing - Original Draft, Review \& Editing; Visualization; Methodology; Programming; Formal Analysis}

\author[2]{Yuejun Yan}
\credit{Project Administration}

\author[1]{Guangyu Wu}
\credit{Writing - Review \& Editing; Methodology}

\author[1]{Siew-Chien Wong}
\credit{Programming; Visualization}

\author[1]{Jimin Jia}
\credit{Writing - Review \& Editing}

\author[2]{Zhaoyang Wang}
\credit{Project Administration}

\author[1]{Yonggang Wen}
\cormark[1]
\ead{ygwen@ntu.edu.sg}
\credit{Supervision; Funding Acquisition; Resources;  Conceptualization}

\affiliation[1]{organization={College of Computing and Data Science},
                addressline={Nanyang Technological University},
                postcode={639798},
                country={Singapore}}
\affiliation[2]{organization={Alibaba Group},
                addressline={Hangzhou},
                postcode={311121},
                country={China}}
                
\cortext[cor1]{Corresponding author}

\begin{abstract}
The growing scale and complexity of modern data centers present major challenges in balancing energy efficiency with outage risk. Although Deep Reinforcement Learning (DRL) shows strong potential for intelligent control, its deployment in mission-critical systems is limited by data scarcity and the lack of real-time pre-evaluation mechanisms. This paper introduces the Dual-Loop Control Framework (DLCF), a digital twin–based architecture designed to overcome these challenges. The framework comprises three core entities: the physical system, a digital twin, and a policy reservoir of diverse DRL agents. These components interact through a dual-loop mechanism involving real-time data acquisition, data assimilation, DRL policy training, pre-evaluation, and expert verification. Theoretical analysis demonstrates how DLCF improves sample efficiency, generalization, safety, and optimality. Leveraging the DLCF, we implemented the DCVerse platform and validated it through case studies on a real-world data center cooling system. The evaluation shows that our approach achieves up to 4.09\% energy savings over conventional control strategies, without violating SLA requirements. Additionally, the framework improves policy interpretability and supports more trustworthy DRL deployment. This work provides a foundation for reliable AI-based control in data centers and points toward future extensions for holistic, system-wide optimization.
\end{abstract}



\maketitle
\section{Introduction}
Rising data center demand is driving both power consumption and system complexity. Driven by the explosive growth of Artificial Intelligence (AI), big data, cloud computing, and digital transformation, global data center capacity demand is projected to increase by 22\% annually through 2030, with advanced AI workloads alone growing at an annual rate of 33\%~\cite{mckinsey2024aipower}. This surge in computational requirements is significantly increasing electricity consumption: data centers accounted for 3\% of global electricity use in 2017, and this figure is expected to rise to 8\% by 2030~\cite{iea2024}. Meanwhile, the scale of data center systems is also increasing in response to this rising capacity demand. For instance, some mega data centers now exceed one million square feet in area and house tens of thousands of servers. This rapid expansion introduces new layers of system complexity, 
including hardware heterogeneity, scalability limitations, and tighter integration requirements across subsystems \cite{kant2009data}.

\textcolor{black}{Rising energy consumption and increasing system complexity are creating Operation and Maintenance (O\&M) challenges for modern data centers that exceed the capabilities of traditional practices.} Managing cost has become a critical concern, while sustainability goals are increasingly shaped by regulatory mandates and societal expectations~\cite{uptime2024survey}. These concerns are commonly evaluated using energy efficiency metrics such as Power Usage Effectiveness (PUE). Nevertheless, after years of improvement, the industry-average PUE has plateaued, signaling diminishing returns from conventional optimization methods and underscoring the need for new technologies to achieve further efficiency gains. Concurrently, the risk of costly outages remains significant, with nearly 20\% of high-impact incidents resulting in financial losses exceeding \$1 million. As a result, operators are under growing pressure to enhance efficiency while ensuring uninterrupted and safe operations~\cite{Uptime2023Survey}, often leading to conservative design and management strategies. Thus, balancing energy efficiency improvements with outage risk mitigation requires the integration of more advanced O\&M strategies~\cite{avelar2023ai}.

\textcolor{black}{Operators are increasingly turning to AI for smarter O\&M.} Industry surveys indicate that AI adoption in O\&M is primarily motivated by its potential to enhance facility efficiency, reduce human error, improve staff productivity, and lower the risk of equipment failure or outages~\cite{uptime2024survey}. These benefits have driven the integration of AI-based capabilities into data center management software, with cooling optimization emerging as a key application area. Among various AI approaches, DRL has gained attention due to its ability to learn adaptive control policies in complex and uncertain environments~\cite{zhang2021survey, kahil2025reinforcement}. In the context of data center cooling, DRL algorithms have been studied for optimizing the operation of chillers, Computer Room Air Handler (CRAH) units, and localized cooling devices~\cite{zhang2023deep}, demonstrating promising results in simulation-based evaluations~\cite{le2021deep, li2019transforming, ran2022optimizing, qingang2023intelligent, wan2023safecool}. Similarly, in the Information and Communication Technology (ICT) subsystem, DRL has been explored for task scheduling and dynamic resource allocation~\cite{yi2020efficient, ghasemi2024energy, ghasemi2024enhancing}. Recent efforts further investigated the joint optimization of cooling and ICT subsystems, aiming to enhance global efficiency and reduce localized thermal hotspots~\cite{zhou2021joint}. 

\textcolor{black}{Despite the growing interest in AI, several critical challenges hinder its widespread deployment.} Survey results show a consistent decline in trust toward AI-based decision-making over the past three years, with the proportion of respondents expressing confidence dropping from 76\% in 2022 to 58\% in 2024 ~\cite{uptime2024survey}. This decline is largely attributed to increasing operator awareness of AI’s limitations, particularly its potential to introduce new points of failure and the lack of transparency in decision-making. These concerns align with the broader trajectory of AI adoption in data centers as characterized by the Hype Cycle~\cite{oosterhoff2020artificial}, where the technology appears to be entering the “trough of disillusionment” following an initial phase of inflated expectations. In response, researchers have emphasized the necessity to revisit existing deployment frameworks and develop new approaches that foster operator trust in AI-driven optimization. Drawing from operator feedback and a synthesis of recent research \cite{zhang2023deep, kahil2025reinforcement}, it can be observed that existing frameworks often fail to deliver well-trained AI models at the point of deployment. In addition, the lack of real-time pre-evaluation mechanisms hinders the ability to assess policy behavior in advance, compromising transparency and limiting confidence in AI-enabled optimization.

\textcolor{black}{These gaps motivate the development of a digital twin–enabled control framework that embeds domain knowledge and enables real-time pre-evaluation, i.e., DLCF.} In this framework, a high-fidelity digital twin serves as the core of the cyber loop, continuously updated through data assimilation from real-time sensing in the physical loop. The digital twin leverages hybrid modeling by integrating physics-based and data-driven techniques, enabling accurate and computationally efficient dynamic prediction. The digital twin supports the pre-training of diverse DRL policies, enables risk-free exploration and rapid adaptation, and facilitates the rigorous pre-evaluation of candidate policies. Therefore, DLCF addresses the key limitations of existing DRL deployment paradigms. \textcolor{black}{The key contributions of this paper are as follows:
\begin{itemize}
    \item We propose DLCF, a framework that uses digital twins for reliable DRL algorithm deployment. DLCF features hybrid digital twin modeling combining physics-based and data-driven methods, a DRL policy reservoir, real-time updates through data assimilation, and a built-in mechanism for real-time policy pre-evaluation.
    \item We present theoretical analyses to elucidate the fundamental attributes of the proposed framework, encompassing its mechanisms for augmenting sample efficiency, generalization, safety, and optimality. We demonstrate how the integration of a high-fidelity digital twin enhances the reliability of DRL.
    \item We implemented the DCVerse platform based on the DLCF framework and validated it through case studies on a real-world data center cooling system. The results demonstrate notable energy savings compared to existing best practices, while maintaining full compliance with SLA requirements and enhancing policy interpretability through pre-evaluation within the digital twin.
\end{itemize}
}

The remainder of this paper is organized as follows. Section~\ref{sec:related_work} reviews related works on DRL-based optimization in data centers and the application of digital twins for DRL control. Section~\ref{sec:framework} describes the proposed DLCF, detailing its architecture and workflow. Section~\ref{Theoretical Analysis} details the theoretical justifications. Section~\ref{sec:experiments} presents the experimental setup and case study results for applying the framework to data center cooling, followed by a discussion of future research directions. Finally, Section~\ref{sec:conclusion} concludes the paper.

\section{Related Works}
\label{sec:related_work}
To contextualize our contributions, we first review related efforts in applying DRL to data center control tasks, followed by a survey of recent advancements in leveraging digital twins to enhance the reliability, safety, and efficiency of DRL deployments. This section highlights the evolution of key ideas and identifies current research gaps that motivate our proposed framework.

\subsection{DRL Control in Data Centers}
Early investigations into the use of DRL in data center systems primarily examined its feasibility for enhancing thermal and energy efficiency. These efforts were commonly conducted in simulated testbeds, aiming to explore how DRL algorithms could coordinate control strategies for both cooling infrastructure and ICT workloads. For example, Li et al.~\cite{li2019transforming} implemented a Deep Deterministic Policy Gradient (DDPG) algorithm to tune operating setpoints for direct expansion and chilled water (CHW) cooling systems, targeting reduced energy usage while maintaining thermal compliance. Leveraging EnergyPlus simulations along with empirical traces from the NSCC in Singapore, this approach reported energy savings of up to 15\%. Mahbod et al.~\cite{mahbod2022energy} deployed the Soft Actor-Critic (SAC) method to manage cooling in a tropical facility, achieving 5.5\% savings in part-load and 3\% in full-load scenarios. Yi et al.~\cite{yi2020efficient} developed a Deep Q-Network (DQN) based job scheduler, trained offline using LSTM-enhanced thermal and power models, which demonstrated over 10\% power reduction in a 1,152-core data center without sacrificing computational throughput. While these early works were largely limited to simplified simulation environments, they laid a foundation for further advances in DRL-enhanced cooling and resource management.

As the field evolved, attention shifted toward addressing real-world deployment challenges of DRL in operational data centers. For instance, Zhang et al.~\cite{zhang2022residual, zhang2023drl}, Wang et al.~\cite{wang2024green}, Le et al.~\cite{le2021deep}, and Cao et al.~\cite{cao2023toward} examined safety concerns arising during policy deployment in dynamic environments. To address both soft and hard constraints, various methods were introduced, including constrained Markov Decision Processes (MDPs), post-hoc action filtering, and Lyapunov-based techniques for stability assurance. In a parallel direction, Zhang et al.~\cite{zhang2024uncertainty} proposed a learning-augmented Model Predictive Control (MPC) scheme that incorporates uncertainty modeling via Monte Carlo trajectory sampling to enable safe exploration during online learning. To improve generalizability and reduce training overhead, Zhang et al.~\cite{zhang2022residual} and Wang et al.~\cite{wang2024green} further explored the integration of domain knowledge into the policy learning process. In response to the data inefficiency of standard DRL approaches, model-based methods have emerged, such as model-based DRL and learning-enhanced MPC~\cite{lazic2018data, wan2023safecool, mu2024large, zhang2024uncertainty}, offering a balance between exploration cost and safety. Moreover, offline DRL techniques trained on historical logs have also gained traction as an alternative for risk-averse deployment~\cite{zhan2025data}. Despite these promising developments, the application of DRL in production-scale data centers remains in its early stages.
\begin{figure}[!t]
    \centering
    \includegraphics[width=0.6\textwidth]{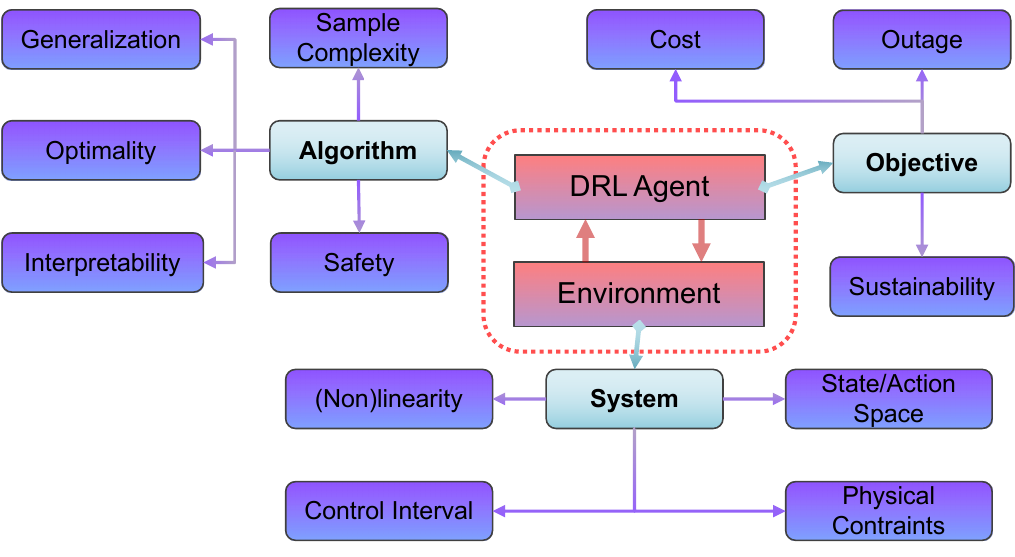}
    \caption{Three key aspects should be considered in algorithm design: system characteristics, optimization objectives, and algorithmic properties.}
    \label{DRLAgent}
\end{figure}

\begin{figure}[!t]
    \centering
    \includegraphics[width=0.65\textwidth]{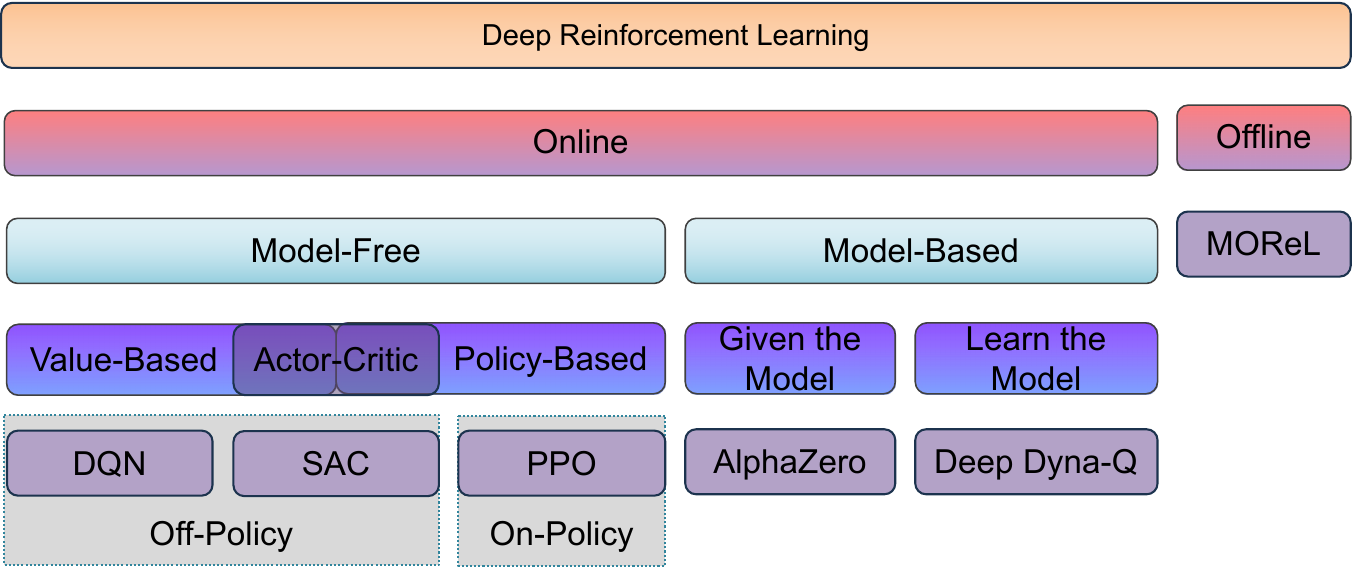}
    \caption{A representative taxonomy of deep reinforcement learning algorithms.}
    \label{DRLTaxonomy}
\end{figure}

Recent studies revisited the challenge of reliably deploying DRL in data centers~\cite{qingang2023intelligent, zhang2023deep, kahil2025reinforcement}. As illustrated in Fig.~\ref{DRLAgent}, DRL design should consider three key aspects: system characteristics, optimization objectives, and algorithmic properties. System characteristics refer to the physical and operational features of the environment, such as (non)linearity, control interval, state and action space dimensionality, and physical constraints. Optimization objectives define the high-level goals of the control task, including cost reduction, outage prevention, and sustainability. Regarding DRL algorithm, Fig.~\ref{DRLTaxonomy} presents a representative taxonomy, including offline DRL, online model-based DRL, and online model-free DRL~\cite{kahil2025reinforcement, prudencio2023survey}. Each category corresponds to a different learning framework and exhibits distinct algorithmic properties in terms of generalization, sample complexity, optimality, safety, and interpretability. Importantly, algorithm design should be aligned with both system characteristics and optimization goals. For instance, systems with long control intervals and complex physical constraints impose strong demands on sample efficiency, making model-based or offline DRL algorithms more suitable. 

Addressing the deployment challenges of DRL requires a higher-level design paradigm that accounts for domain-specific factors, such as data scarcity and the need for real-time pre-evaluation. In this context, frameworks that integrate digital twins offer a promising direction.

\subsection{Digital Twins for DRL Control}
Digital twins have gained increasing traction across various domains for improving system observability and control. As defined by the National Academies of Sciences, Engineering, and Medicine~\cite{national2023foundational}, a digital twin refers to a virtual representation that replicates the structure, operating context, and dynamic behavior of a physical asset. A fundamental aspect of this concept is the continuous bidirectional exchange of information between the physical system and its digital counterpart. While originally rooted in aerospace and manufacturing for lifecycle management, the application of digital twins has since expanded to include smart grids, autonomous systems, and data center infrastructures~\cite{tao2022digital, faraboschi2023digital,zhang2025caper}.

Coupling digital twins with DRL has emerged as a powerful strategy for intelligent system control. For instance, Schena et al.~\cite{schena2024reinforcement} introduced a reinforcement twinning architecture that jointly leverages model-based and model-free RL within a digital twin environment. Their method trains both policies concurrently and enables role-switching based on observed performance. Validated across multiple domains—including wind turbine regulation, UAV trajectory planning, and cryogenic tank operations—the framework achieved gains in sample efficiency and control robustness. In the networking domain, Zhang et al.~\cite{zhang2024digital} developed a DRL framework enhanced by digital twins to address network slicing challenges. Their approach constructs a data-driven twin from historical measurements to emulate real-time dynamics, with the DRL agent trained within the virtual environment for risk-averse decision-making. Similarly, Cheng et al.~\cite{cheng2024toward} proposed a twin-assisted DRL solution for managing network resources. Experiments in ultra-reliable low-latency communication and UAV coordination confirmed improvements in both convergence speed and policy quality. The integration of DRL and digital twin technologies has also extended to manufacturing~\cite{xia2021digital}, smart grid optimization~\cite{zhou2023digital}, and electric mobility applications~\cite{ye2024deep}. Specifically in the data center context, Athavale et al.~\cite{athavale2024digital} presented a holistic digital twin vision, supporting optimization tasks such as thermal management, energy saving, workload scheduling, failure anticipation, and carbon reduction.

These studies represent promising initial efforts to integrate digital twins with DRL. Nevertheless, a unified framework is still lacking to systematically characterize the interactions among physical systems, digital twins, and DRL agents. In particular, the theoretical understanding of how digital twins contribute to improving the performance of DRL remains limited.

\section{Methodology}
\label{sec:framework}
In this section, we first introduce foundational concepts in DRL. We then present the proposed DLCF, detailing its three core entities—the physical system, the digital twin, and the policy reservoir—and the interactions among them through a dual-loop architecture. 

\subsection{Reinforcement Learning Preliminaries}
RL is typically framed as a MDP, defined by the tuple \(\langle \mathcal{S}, \mathcal{A}, M, R, \gamma \rangle\), where \(\mathcal{S}\) denotes the state space, \(\mathcal{A}\) represents the action space for the agent, \(M(s'\mid s,a)\) indicates the probability of reaching state \(s'\) after taking action \(a\) in state \(s\), \(R(s,a)\) is the reward function providing immediate rewards for action \(a\) in state \(s\), and \(\gamma \in [0,1)\) is the discount factor assessing the importance of future rewards.

A policy $\pi$ defines the agent's behavior, specifying the probability of taking action $a$ in state $s$:
\begin{equation}
\label{policy_function}
    \pi(a \mid s) = P(A_t = a \mid S_t = s).
\end{equation}
The objective of RL is to find a policy \(\pi\) that maximizes the expected cumulative discounted reward:
\begin{equation}
\label{objective_function}
    J(\pi) = \mathbb{E}_{\tau \sim \pi}\left[\sum_{t=0}^{\infty} \gamma^t R(S_t, A_t)\right],
\end{equation}
where a trajectory \(\tau = (S_0, A_0, S_1, A_1, \dots)\) follows policy \(\pi\). The optimal policy \(\pi^*\) maximizes this objective:
\begin{equation}
    \pi^* = \arg\max_{\pi} J(\pi).
\end{equation}

Value functions estimate the performance of a policy. The state-value function $V^\pi(s)$ is defined as:
\begin{equation}
\label{state_value_function}
    V^\pi(s) = \mathbb{E}\left[\sum_{t=0}^{\infty}\gamma^t R(S_t,A_t) \mid S_0 = s\right],
\end{equation}
which represents the expected return starting from state $s$ under policy $\pi$. The state-action value function (Q-function) is defined as:
\begin{equation}
    Q^\pi(s,a) = \mathbb{E}\left[\sum_{t=0}^{\infty}\gamma^t R(S_t,A_t) \mid S_0=s, A_0=a\right],
\end{equation}
representing the expected return starting from state $s$, taking action $a$, and following policy $\pi$ thereafter.

In Model-Based DRL, an agent learns approximations of the transition dynamics \(M(s'|s,a)\) and reward function \(R(s,a)\). The transition dynamics can be represented either deterministically or stochastically. A deterministic model parameterized by \( \theta \) can typically be trained by minimizing a loss function such as Mean Squared Error:
\begin{equation}
\label{deterministic_transition}
\min_{\theta} \mathbb{E}_{(s,a) \sim \rho_{\pi},\; s' \sim M(\cdot|s,a)} \left[\| s' - \hat{M}_{\theta}(s,a)\|_2^2\right],
\end{equation}
\(\rho_{\pi}\) represents the empirical state-action distribution from the data collection policy \(\pi\). To model dynamics uncertainty, a stochastic model minimizes the divergence between the true distribution \(M(\cdot|s,a)\) and the model's predictive distribution \(\hat{M}_{\theta}(\cdot|s,a)\), using methods like KL divergence,
\begin{equation}
\label{stochastic_transition}
\min_{\theta}\mathbb{E}_{(s,a)\sim\rho_{\pi}}\left[D_{KL}\left(M(\cdot|s,a)\;||\;\hat{M}_{\theta}(\cdot|s,a)\right)\right].
\end{equation}

DRL problems can also be formulated in a finite-horizon episodic setting without discounting, represented as \( \langle \mathcal{S}, \mathcal{A}, M, r, H \rangle \), where \( H \) is the planning horizon. This formulation suits scenarios such as learning-based MPC, where an agent repeatedly solves finite-horizon optimization problems:
\begin{equation}
\max_{\pi} \mathbb{E}_{\tau \sim \pi}\left[\sum_{t=0}^{H-1} R(S_t,A_t)\right].
\end{equation}

\begin{figure*}[!t]
    \centering
    \includegraphics[width=0.8\textwidth]{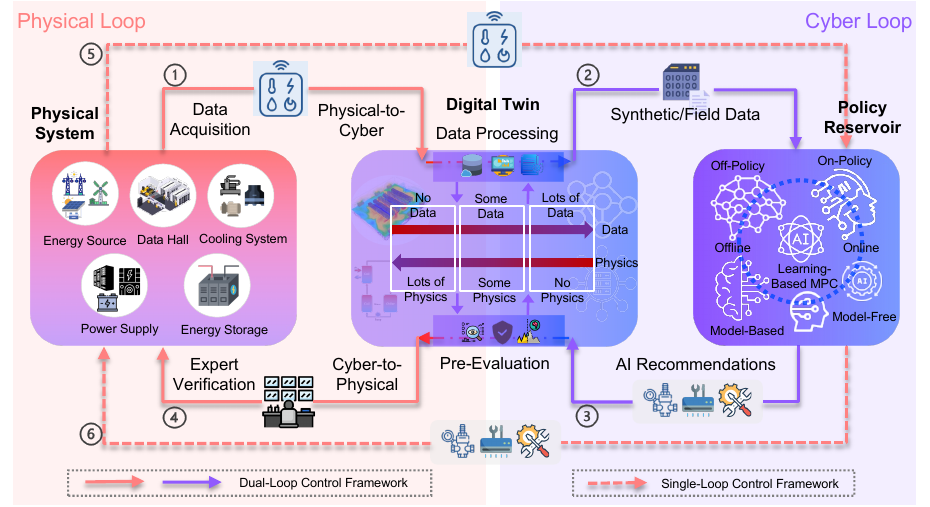}
    \caption{Dual-loop control framework for the DRL deployment. Dual-loop control framework: (1)-(4). Single-loop control framework: (5)-(6).}
    \label{DLCF_Framework}
\end{figure*}

\subsection{Core Entities of the Framework}
The proposed DLCF comprises three interconnected entities: the physical system, the digital twin, and the policy reservoir, as illustrated in Fig.~\ref{DLCF_Framework}.

\subsubsection{Physical System}

The physical system comprises mission-critical data center infrastructure, encompassing IT equipment (e.g., servers, switches, storage units), thermal management components (e.g., chillers, cooling towers, CRAHs), power distribution systems (e.g., uninterruptible power supply, transformers, generators), and integrated renewable energy and storage solutions. These systems typically exhibit complex nonlinear dynamics with high-dimensional state and action spaces due to numerous control inputs (e.g., fan speeds, valve positions, workload placement, power allocations) and operational states (e.g., server load, temperature distribution, relative humidity, air and water flowrate, power flows). The sampling and control intervals are generally on the order of minutes, reflecting system response times and sensor reporting frequencies. Additionally, the physical system is subject to various constraints, including strict physical limitations (e.g., temperature and humidity thresholds) and equipment operational limits (e.g., cooling capacity). High-fidelity IoT sensors equipped with edge-computing capabilities continuously monitor critical parameters, perform local preprocessing (e.g., calibration, noise filtering), and securely transmit refined data to centralized storage, providing a robust data foundation for digital twin modeling and decision-making within the DLCF.

\subsubsection{Digital Twin}

At the core of the DLCF lies a digital twin that functions as a virtual replica of the physical system. Given the dual requirements of accurate dynamic prediction and real-time model updating, the digital twin must simultaneously deliver high computational efficiency and strong generalization capability. To meet these demands, hybrid modeling approaches are employed, combining physics-based and data-driven methods to exploit their complementary strengths.

In the context of data center thermal management, physics-based models—governed by thermodynamics, fluid dynamics, and heat transfer—can accurately capture airflow and temperature distributions via Computational Fluid Dynamics simulations. However, such models are often computationally prohibitive for real-time control. In contrast, data-driven models offer faster inference but typically require large training datasets and exhibit limited generalization beyond observed conditions. To address these limitations, Physics-Informed Machine Learning (PIML) has emerged as a promising approach that embeds physical laws directly into machine learning models. A typical PIML training objective is formulated as:
\begin{equation}
    \mathcal{L}_{\text{PIML}}(\theta) = \mathcal{L}_{\text{data}}(\theta) + \lambda\,\mathcal{L}_{\text{physics}}(\theta),
\end{equation}
where \(\lambda\) controls the trade-off between data-driven accuracy and physical consistency. Beyond thermal management, this hybrid modeling approach is also applicable to other subsystems—such as power distribution, energy storage, and renewable integration—thereby improving the overall accuracy and reliability of the digital twin. These enhancements establish a more robust foundation for DRL training, evaluation, and adaptation within the DLCF framework.

\subsubsection{Policy Reservoir}

The policy reservoir serves as a centralized module that stores a diverse collection of DRL agents spanning multiple algorithmic categories, including offline DRL, online model-based DRL, and online model-free DRL approaches. Each policy is annotated with metadata describing its algorithmic type, applicable operating conditions, target objectives, and historical performance. This metadata facilitates efficient indexing, filtering, and selection of candidate strategies in response to real-time system states. Furthermore, the reservoir supports complementary deployment of policies—for example, combining long-horizon with short-horizon policies, or pairing primary controllers with conservative backup strategies. By maintaining a variety of policies, the reservoir enables context-aware policy selection, supports fallback mechanisms under uncertainty, and enhances robustness under dynamic operational conditions.

\subsection{Dual-Loop Interaction Mechanism}
The DLCF is structured around two interconnected loops: the physical loop, which captures the interaction between the physical system and the digital twin, and the cyber loop, which governs the interaction between the digital twin and the DRL policy reservoir. The complete workflow is summarized in Algorithm~\ref{alg:dual-loop-framework}. 
\begin{algorithm}[t]
\caption{Dual Loop Control Framework}
\label{alg:dual-loop-framework}
\small
\begin{algorithmic}[1]
\State \textbf{Input:} \parbox[t]{0.85\linewidth}{%
System information includes equipment specifications, physics-based models,
historical datasets, \\ real-time sensor data, a library of reinforcement learning algorithms, and defined optimization objectives}
\State Construct the digital twin using hybrid modeling
\State Calibrate digital twin using historical data
\State Initialize policy reservoir
\For{each DRL algorithm in the set}
    \State Train the algorithm within the digital twin
    \State Tune hyperparameters using digital twin simulations
    \State Add the resulting policy to the policy reservoir
\EndFor
\While{system is running}
    \State Collect real-time sensor data
    \State Perform data assimilation
    \State Generate candidate actions from the policy reservoir
    \For{each candidate action $a_j$}
        \State Evaluate $a_j$ in the digital twin environment
        \If{safety is violated}
            \State Project $a_j$ to safe region
        \EndIf
    \EndFor
    \State Select the best candidate action $a^*$
    \State Optionally verify $a^*$ with expert feedback
    \State Deploy $a^*$ to the physical system
\EndWhile
\end{algorithmic}
\normalsize
\end{algorithm}

\subsubsection{Data Acquisition and Assimilation (Flow 1)}

The physical loop begins with the continuous acquisition of real-time data from high-resolution IoT sensors deployed throughout the data center. Once transmitted to the digital twin, the data undergoes additional preprocessing (e.g., outlier detection, missing value imputation, normalization) before being fed into the digital twin models. 

A distinctive advantage of the digital twin is its ability to continuously assimilate real-time data, maintaining accurate representations of the physical system. Data assimilation typically includes state estimation and system identification. System identification, or model parameter calibration when the model structure is fixed, involves continuously updating the model's parameters. This process is commonly formulated as an optimization problem:
\begin{equation}
    \hat{\theta} = \arg\min_{\theta}\sum_{i=1}^{N}\left\|y_i - \hat{y}(x_i,\theta)\right\|_2^2,
\end{equation}
where \(y_i\) are observed measurements, \(\hat{y}\) is model prediction, and \(\theta\) is the model parameter.  State estimation integrates real-time observations to infer current and future system states, extending insight beyond direct sensor measurements. Algorithms such as the Kalman filter and particle filter are selected based on system linearity and complexity. 

Integrating state estimation and system identification within the data assimilation framework ensures ongoing synchronization between digital twins and the physical system. This supports improvements in model fidelity and DRL policy effectiveness, enabling adaptation to system changes and component aging.

\subsubsection{Policy Training and Evaluation (Flow 2-3)}

Within the DLCF framework, the digital twin plays a central role in supporting the training and evaluation of DRL policies. Leveraging its hybrid modeling architecture, the digital twin enables fast yet physically consistent simulations that significantly accelerate the training process. Compared to real-world interactions, the simulated environment allows for risk-free exploration and systematic hyperparameter optimization, leading to the development of more robust and generalizable control strategies. In addition, the digital twin supports parallel training of multiple DRL agents across varied state spaces, control actions, and reward functions, which contributes to policy diversity and specialization. During deployment, the framework dynamically queries the reservoir to retrieve candidate policies based on real-time system conditions.

Before any control action is applied to the physical system, the selected policy undergoes a digital pre-evaluation process (flow 3) within the twin. This evaluation involves simulating the candidate actions under current or forecasted system states and verifying compliance with SLAs and energy efficiency. Actions that fail to meet the predefined criteria are filtered out and projected into the safe region. This evaluation pipeline improves both the reliability and interpretability of DRL-based decision-making and supports safe real-world deployment.

\subsubsection{Expert Verification (Flow 4)}

To further enhance reliability and accountability in DRL-based decision-making, the DLCF incorporates an expert-in-the-loop verification mechanism as a safeguard before control actions are physically executed. While the digital twin’s pre-evaluation ensures that candidate actions meet quantitative performance and safety criteria, certain scenarios still require human judgment, particularly when addressing complex trade-offs beyond the scope of predefined reward functions. In this stage, domain experts—such as facility operators or control engineers—review simulation outcomes from the digital twin and assess whether the recommended actions align with operational experience. When necessary, they may approve the action, modify it, or fall back on a conservative strategy from the policy reservoir. This human-in-the-loop mechanism improves transparency, builds trust in AI-driven control decisions, and supports the safe deployment of autonomous systems in mission-critical environments. Expert feedback can also be logged to refine the pre-evaluation process or guide future policy updates.

\section{Theoretical Analysis}
\label{Theoretical Analysis}
In this section, we provide theoretical perspectives and summarize existing findings to justify the design choices of the proposed framework. We present theoretical analyses to elucidate the mechanisms by which the framework augments sample efficiency, generalization, safety, and optimality. The key results of the analysis are summarized in Table~\ref{tab:comparison}. Interpretability will be further illustrated in the case study presented in Section~\ref{sec:experiments}. Domain independence refers to the extent to which a method operates without relying on domain-specific knowledge.
\begin{table*}[t]
\centering
\caption{Comparison of DLCF and SLCF across key aspects.}
\renewcommand{\arraystretch}{1.3}
\begin{tabular*}{\textwidth}{@{\extracolsep{\fill}} llllll}
\toprule
& DLCF & \multicolumn{2}{l}{SLCF} \\
\cmidrule(lr){3-6}
& & Online model-based & Online model-free & Offline model-based & Vanilla offline \\
\midrule
Sample Efficiency & High & Medium & Low & - & - \\
Generalization    & High & - & - & Medium & Low \\
Optimality        & High & Low & High & Low & Low \\
Safety            & High & Low & Low & Low & Low \\
Interpretability  & High & Low & Low & Low & Low \\
Domain Independence   & Low  & High & High & High & High \\
\bottomrule
\end{tabular*}
\vspace{0.5em}
\label{tab:comparison}
\end{table*}

\subsection{Sample Complexity}
Unlike simulations, real-world interactions in mission-critical systems like data centers are costly—not only in energy and risk of service disruption but also in time, as system responses are often slow. This makes minimizing the number of interactions (i.e., sample complexity) essential. In RL, sample complexity refers to the number of interactions an algorithm requires with the environment to learn an $\epsilon$-optimal policy with probability at least $1 - \delta$ \cite{kakade2003sample}. That is, the number of interactions needed to ensure that the learned policy $\pi$ satisfies:
\begin{align}
    V^{\pi}(s) &\ge V^*(s) - \epsilon, \nonumber \\
    &\text{with probability at least } 1 - \delta,\quad \forall s \in \mathcal{S}
\end{align}
where $\epsilon$ is the accuracy tolerance (optimality gap), $\delta$ is the failure probability bound, $\mathcal{S}$ is the state space. $\pi$ and $V^{\pi}(s)$ are defined in Eq.~(\ref{policy_function}) and Eq.~(\ref{state_value_function}), respectively. The number of samples typically refers to the number of time steps or $(s,a,a')$ transitions from the real-world system. Despite having a sufficiently expressive function class capable of representing the optimal Q-function, model-free RL still suffers from fundamental sample inefficiency. Sun et al.~\cite{sun2019model} establish a lower bound showing that even with realizability, i.e., when the optimal Q-function lies within the function class \( \mathcal{G} = \mathrm{OP}(\mathcal{M}) \), where \( \mathrm{OP}(\mathcal{M}) \) denotes the set of optimal Q-functions and policies induced by all models in a realizable model class \( \mathcal{M} \) (i.e., \( \mathrm{OP}(M) \triangleq (Q_M, \pi_M) \)), any model-free algorithm must collect at least \( \Omega(2^H) \) trajectories to learn a near-optimal policy. Otherwise, with high probability (at least \( 1/3 \)), the algorithm will output a policy \( \hat{\pi} \) whose value is worse than the optimal by a constant margin: $v^{\hat{\pi}} < v^* - 1/2$. This exponential dependence on the planning horizon \( H \) highlights a fundamental limitation of model-free methods: even when function approximation is not the bottleneck, the lack of structural modeling leads to prohibitive exploration requirements in complex environments.

In model-based RL, the sample complexity is tightly linked to the structural properties of the model class. Sun et al.~\cite{sun2019model} introduce a structural complexity measure called the witness rank $W_\kappa$, which quantifies the number of independent test functions required to distinguish between candidate models within a given test function class $\mathcal{F}$. Under standard assumptions—including realizability (i.e., the true model $M^* \in \mathcal{M}$) and Bellman domination (which ensures the Bellman error can be bounded by model misfit)—they derive a sample complexity upper bound for learning a near-optimal policy:
\begin{equation}
\label{model_based_upper_bound}
    N_{\text{MB}} = \tilde{O} \left( \frac{H^3 \cdot W_\kappa^2 \cdot K}{\kappa^2 \epsilon^2} \log \left( \frac{T \cdot |\mathcal{F}| \cdot |\mathcal{M}|}{\delta} \right) \right),
\end{equation}
where $\kappa \in (0,1]$, $T = H \mathcal{W}_\kappa \log(\beta / 2\phi) / \log(5/3)$, and $\phi = \kappa \epsilon / (48 H \sqrt{W_\kappa})$. Here, $K = |\mathcal{A}|$ denotes the cardinality of the action space $\mathcal{A}$. For a detailed derivation and theoretical guarantees, please refer to~\cite{sun2019model}. This bound underscores the critical role of model class structure and the witness rank $W_\kappa$: the more structured and identifiable the model class $\mathcal{M}$ is (i.e., the smaller the witness rank), the fewer samples are required to learn an effective policy. Therefore, model-based methods can achieve significantly higher sample efficiency- often exponentially better- compared to their model-free counterparts.

In a model-based RL framework, the structural complexity of the model hypothesis space \(\mathcal{M}\) fundamentally influences the theoretical efficiency of learning.  Consider an unconstrained model hypothesis space \(\mathcal{M}\). 
Without structural priors, candidate models can differ arbitrarily across a high-dimensional space, leading to large discrepancies captured by the Witnessed Model Misfit~\cite{sun2019model}, defined as:
\begin{equation}
\mathcal{W}(M, M', h; \mathcal{F}) \triangleq\; \sup_{f\in\mathcal{F}} \mathbb{E}_{s_h\sim\pi_M,\, a_h\sim\pi_{M'}} \Bigg[ 
\mathbb{E}_{(r,s')\sim M'_h}\left[f(s_h,a_h,r,s')\right] 
-\; \mathbb{E}_{(r,s')\sim M_h^*}\left[f(s_h,a_h,r,s')\right] 
\Bigg]
\end{equation}
for any models \( M, M' \in \mathcal{M} \) and timestep \( h\in[H] \). $M_h^*$ is the true system dynamics model. The Witness Rank $W_\kappa$ is then defined based on bounding matrices sandwiched between the Witnessed Model Misfit $\mathcal{W}(M, M', h)$ and Bellman error matrices $\mathcal{E}_B(M, M', h)$: $W_{\kappa}(\kappa, \beta, \mathcal{M}, \mathcal{F}, h) = \min_{A\in\mathcal{N}_{\kappa,h}} \text{rank}(A, \beta)$,
where \(\mathcal{N}_{\kappa,h}\) denotes the set of matrices satisfying:
\begin{align}
\kappa \mathcal{E}_B(M, M', h) 
&\leq A(M, M') \notag\\
&\leq \mathcal{W}(M, M', h), \quad \forall M, M' \in \mathcal{M}.
\end{align}
Without structural priors, the differences between candidate models are rich and complex, leading to a high Witness Rank \(W_\kappa\) and, consequently, higher sample complexity.

When physical priors are incorporated, they impose strong structural constraints on the model class.  This results in a much smaller effective hypothesis space \(\mathcal{M}_{\text{phys}} \subset \mathcal{M}\), satisfying:
$
|\mathcal{M}_{\text{phys}}| \ll |\mathcal{M}|.
$
Consequently, the Witness Rank is substantially reduced:
$
\mathcal{W}_{\kappa,\text{phys}} \ll \mathcal{W}_{\kappa}.
$
Thus, according to Eq.~(\ref{model_based_upper_bound}), physical priors substantially reduce the Witness Rank, resulting in provable reductions in sample complexity and enabling more efficient and robust model-based RL.

\subsection{Generalization}
In real-world data centers, operational and safety constraints limit system state exploration. Historical data gathered under routine, rule-based control policies show low diversity and limited state-action space coverage, posing challenges for offline RL. Let \( \mathcal{D} = \{(s_i, a_i, r_i, s_i')\}_{i=1}^N \) be the offline dataset collected under a behavior policy \( \pi_\beta \). The distribution over state-action pairs is denoted \( \rho^{\pi_\beta}(s,a) \). However, the learned policy \( \pi \) induces its own distribution \( \rho^\pi(s,a) \), which may deviate significantly from \( \rho^{\pi_\beta} \). In this setting, estimating the performance of \( \pi \) using a value function \( \hat{Q} \) trained solely on \( \mathcal{D} \) introduces extrapolation error: $\epsilon_{\text{gen}}(\pi) = \mathbb{E}_{(s,a) \sim \rho^\pi} \left[ Q^\pi(s,a) - \hat{Q}(s,a) \right]$. Since \( (s,a) \sim \rho^\pi \) may lie outside the support of \( \mathcal{D} \), this error can be large, leading to poor generalization. This is the core manifestation of the Out-Of-Distribution (OOD) generalization problem in Offline RL.

Model-based offline RL addresses this issue by leveraging a learned dynamics model \( \hat{M}(s'|s,a) \), trained using the same offline dataset \( \mathcal{D} \) \cite{yu2020mopo}. This model is then used to generate synthetic transitions, {\color{black} {namely serves as the mapping $\hat{M} :(s, a) \mapsto \hat{s}'$}},$\quad \hat{r} = \hat{R}(s,a)$, allowing the construction of an extended dataset \( \tilde{\mathcal{D}} \) that better covers the target policy distribution \(\rho^\pi \). The Q-function can then be trained with a Bellman-consistent objective over this model-augmented dataset. In the MOReL framework~\cite{kidambi2020morel}, a pessimistic model \(\hat{M}_p\) is constructed to mitigate model uncertainty during offline RL. For any state-action pair \((s,a)\) deemed uncertain, the model transitions deterministically to an absorbing HALT state with a fixed negative reward penalty, while otherwise following the learned transition \(\hat{M}(s'|s,a)\) and reward \(\hat{R}(s,a)\). This design enforces conservatism in poorly estimated regions, preventing the learned policy from exploiting model inaccuracies. Key notations include \(M^*\) (true MDP), \(\hat{M}\) (learned model), \(\hat{M}_p\) (pessimistic model), \(\rho_0\) (initial state distribution), $\hat{\rho}_0$ (estimated initial state distribution), and \(D_{TV}(\rho_0, \hat{\rho}_0)\) (total variation distance). MOReL establishes performance bounds between the policy value evaluated on the pessimistic model \(\hat{M}_p\) and the true environment \(M^*\). Specifically, for any policy \(\pi\), the following two-sided inequalities hold:
\begin{equation}
\label{lower_bound}
J_{\hat{\rho}_0}(\pi, \hat{M}_p) 
\geq J_{\rho_0}(\pi, M^*) 
- \frac{2R_{\max}}{1-\gamma} D_{TV}(\rho_0, \hat{\rho}_0) 
- \frac{2\gamma R_{\max}}{(1-\gamma)^2} \alpha 
- \frac{2R_{\max}}{1-\gamma} \mathbb{E}\left[\gamma^{T_u^\pi}\right],
\end{equation}
\begin{equation}
\label{upper_bound}
J_{\hat{\rho}_0}(\pi, \hat{M}_p)
\leq J_{\rho_0}(\pi, M^*) 
+ \frac{2R_{\max}}{1-\gamma} D_{TV}(\rho_0, \hat{\rho}_0) 
+ \frac{2\gamma R_{\max}}{(1-\gamma)^2} \alpha,
\end{equation}
where \(D_{TV}(\rho_0, \hat{\rho}_0)\) measures the mismatch between the true and estimated initial state distributions, \(\alpha\) quantifies the maximum model error across state-action pairs, and \(\mathbb{E}\left[\gamma^{T_u^\pi}\right]\) captures the discounted probability of the policy entering the uncertain region.

In model-based offline RL, the maximum local model error \(\alpha\) quantitatively measures the fidelity of the learned transition probabilities and is formally defined as the maximum total variation distance across all state-action pairs:
\begin{equation}
    \alpha = \max_{(s,a)\in\mathcal{S}\times\mathcal{A}} D_\text{TV}\left( \hat{M}(\cdot|s,a), M^*(\cdot|s,a) \right).
\end{equation}
Without structural constraints, the model hypothesis space (e.g., \(\mathcal{M}_\text{DNN}\) defined by Deep Neural Networks, i.e., DNNs) is extremely flexible and typically of high complexity, causing potentially large errors, especially in regions with sparse data coverage. Consequently, the unconstrained maximum local model error can be large.
When explicit physics priors or constraints are incorporated, the estimated transition probabilities are constrained strictly within a physically feasible low-dimensional manifold:
$
    \hat{M}_{\text{phys}}(\cdot|s,a) \in \mathcal{M}_{\text{phys}}, \quad |\mathcal{M}_{\text{phys}}| \ll |\mathcal{M}_\text{DNN}|.
$
In this scenario, differences among candidate models are strictly confined within a low-dimensional subspace permitted by physical laws. 
Such strong dimensionality reduction imposed by physics priors directly leads to a significant decrease in the maximum local model error \(\alpha\):
\begin{equation}
    \alpha_{\text{phys}} = \max_{(s,a)\in\mathcal{S}\times\mathcal{A}} D_\text{TV}\left( \hat{M}_{\text{phys}}(\cdot|s,a), M^*(\cdot|s,a) \right) \ll \alpha_{\text{DNN}}.
\end{equation}

Physics priors reduce $\alpha$, which mathematically tightens the upper and lower bounds in Eq.~(\ref{lower_bound}) and Eq.~(\ref{upper_bound}), resulting in smaller uncertainty and better guarantees for the policy performance. Therefore, incorporating a model is also crucial for offline RL where environment interaction is restricted to enable effective generalization.

\subsection{Algorithmic Diversity}
Policy performance in RL can vary significantly due to factors such as random initialization, exploration strategies, hyperparameter sensitivity, and optimizer behavior. To mitigate this variability, we adopt algorithmic diversity, where multiple RL algorithms—or multiple configurations (e.g., random seeds, architectures, or learning rates) of the same algorithm—are trained independently in the digital twin, resulting in a set of policies \( \Pi = \{\pi_1, \pi_2, \dots, \pi_K\} \).

Instead of committing to a single potentially suboptimal policy, we select the best-performing policy from \( \Pi \) by: (1) offline evaluation on held-out scenarios; (2) adaptive selection based on real-time feedback during deployment; or (3) ensemble-based methods such as voting or policy averaging. Let \( M \) denote a fixed environment. The expected cumulative reward of any policy \( \pi \) is denoted by \( J_M^\pi \). We define $J_{\max}^{\Pi} := \max_{\pi \in \Pi} J_M^\pi$. That is, \( J_{\max}^{\Pi} \) represents the best achievable performance among the policy set \( \Pi \). For any policy \( \pi \in \Pi \), it holds that $J_{\max}^{\Pi} \geq J_M^\pi$.

This result formalizes the intuition that algorithmic diversity guarantees that the selected policy from \( \Pi \) will perform at least as well as any single policy in the set, improving worst-case reliability. It is particularly effective in mitigating performance variability caused by training randomness, leading to more robust RL deployments.

\subsection{Optimality}

In the proposed dual-loop framework, the continuous data assimilation and model updating capability of the digital plays a crucial role in tightening the performance bounds of DRL policies. As the digital twin system accumulates real-time sensor data and systematically updates its hybrid models, the discrepancy between the learned model \(\hat{M}_p\) and the true environment model \(M^*\) progressively decreases. Given a policy \(\pi\), the lower bound and upper bound are indicated in Eq.~(\ref{lower_bound}) and Eq.~(\ref{upper_bound}).

As the digital twin is continuously updated with new sensor measurements and system feedback, the following effects are observed:
\begin{itemize}
    \item The distributional error \(D_{TV}(\rho_0, \hat{\rho}_0)\) progressively shrinks. This effect is supported by classical results in statistical learning theory. As more sensor observations are collected, empirical estimates of the initial state distribution converge to the true distribution.
    \item The expected long-horizon modeling error $\alpha$ diminishes. As the one-step prediction error decreases with accumulating data, the compounded error over multi-step rollouts diminishes accordingly.
    \item As the digital twin assimilates more data and improves its modeling fidelity, the uncertain region within the state-action space progressively shrinks. Consequently, the expected discounted probability \(\mathbb{E}\left[\gamma^{T_u^\pi}\right]\) of a policy encountering regions with significant model errors decreases over time.
\end{itemize}
Formally, in the idealized limit where the digital twin model becomes an exact replica of the physical system (i.e., \(\hat{M}_p \to M^*\) and \(\hat{\rho}_0 \to \rho_0\)), we have:
\begin{equation}
\lim_{\hat{M}_p \to M^*,\, \hat{\rho}_0 \to \rho_0} 
\left| J_{\hat{\rho}_0}(\pi, \hat{M}_p) - J_{\rho_0}(\pi, M^*) \right| = 0.
\end{equation}

Thus, continuous model refinement through the digital twin fundamentally tightens the theoretical performance bounds, reduces uncertainty in policy outcomes, and facilitates the deployment of safer and more reliable DRL agents in real-world environments.

\subsection{Safety}

Before deploying RL policies to real-world systems, it is crucial to ensure that policy behaviors comply with hard safety constraints (e.g., temperature, power limits). Directly deploying unverified policies may cause safety hazards or equipment damage. To address this, we propose a digital twin-based pre-validation mechanism that simulates and filters unsafe behaviors before deployment.

We consider safety constraints of the form: the policy must satisfy \( C(s_t, a_t) \leq \epsilon_c \) at all times, where \( C(s, a) \) is a constraint function and \( \epsilon_c \) is the system safety threshold. The standard RL objective is to maximize the expected return in Eq.~(\ref{objective_function}), which does not guarantee constraint satisfaction. Therefore, we introduce two mechanisms: action projection and pre-evaluation before deployment using a digital twin. The policy \( \pi \) outputs an action \( a \notin \mathcal{A}_\text{safe} \), we apply a projection:
\begin{equation}
    a_{\text{safe}} = \arg\min_{a' \in \mathcal{A}_\text{safe}} \| a' - a \|_2,
\end{equation}
which maps the original action to the nearest admissible action in the constrained set. The resulting safe policy is defined as $\pi_{\text{safe}}(a \mid s)$.

However, such runtime projection may interfere with optimality because the projected policy may differ from the original, potentially degrading the return. Therefore, we advocate for safer alternatives by verifying policy candidates in simulation before deployment. Let \(\Pi_{\text{candidate}} = \{ \pi_1, \pi_2, \dots, \pi_K \}\) denote a set of projected safe policy candidates obtained through training, search, or imitation learning. A digital twin constructs an environment model \(\hat{M}\) to evaluate the expected performance of each policy under safety constraints. Formally, the constrained policy selection problem is defined as:
\begin{equation}
\pi^* = \arg\max_{\pi \in \Pi_{\text{candidate}}} \; \mathbb{E}_{s_0 \sim \rho_0,\, \tau \sim (\pi, \hat{M})} \left[ R(\tau) \right],
\end{equation}
where \(R(\tau)\) denotes the cumulative reward along a trajectory \(\tau\), \(s_0\) is the current state, and \(\tau\) is generated by executing \(\pi\) in the model \(\hat{M}\).

\section{Case Study}
\label{sec:experiments}
We incorporate the DLCF into our DCVerse system, which is an integrated platform designed to advance data center operations using digital twins and AI \cite{imda2025reddot}. We then conduct case studies based on a digital twin of a real-world data center. DRL agents are trained to regulate the cooling system using the proposed dual-loop framework. The digital twin is further utilized to analyze agent behavior, thereby enhancing the interpretability of the learned policies and providing deeper insight into their decision-making mechanisms.

\subsection{Real-World System Setup}
\begin{figure}[!t]
    \centering
    \includegraphics[width=0.7\textwidth]{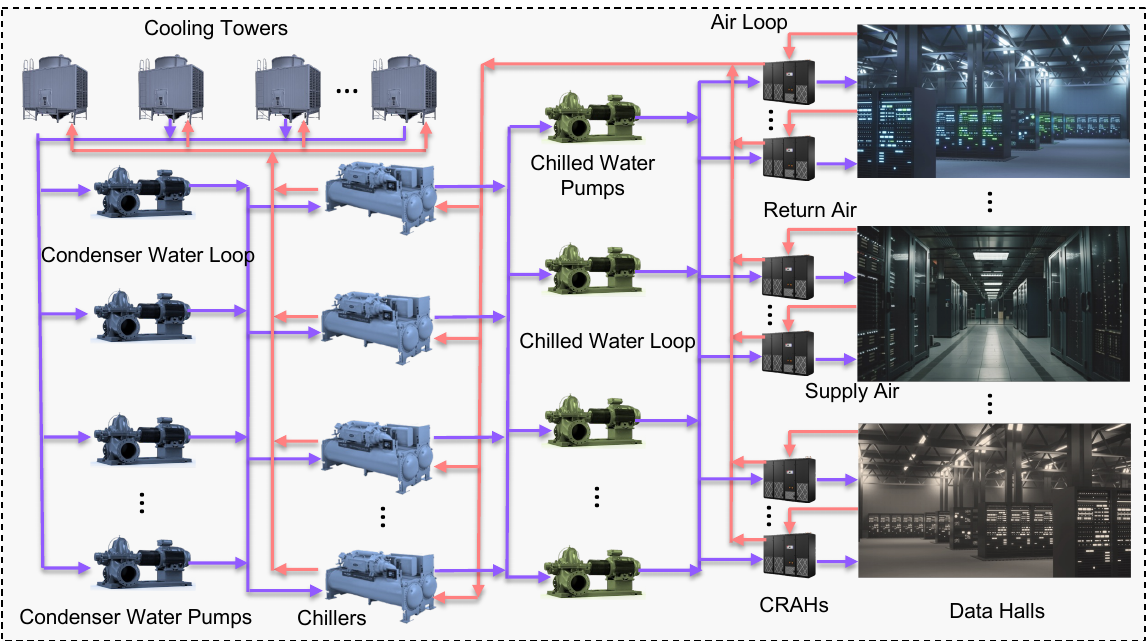}
    \caption{Real-world data center cooling layout comprising three loops: condenser water loop, chilled water loop, and air loop.}
    \label{DataCenterLayout}
\end{figure}

The developed high-fidelity digital twin is configured based on a practical data center deployment, as depicted in Fig.~\ref{DataCenterLayout}. An aisle containment strategy is employed to manage airflow patterns. The DRL agent is designed to minimize the overall energy consumption of the cooling system, encompassing components such as CRAH units, chilled water pumps, chillers, condenser pumps, and cooling towers. Control variables manipulated by the agent include the chilled water supply temperature, the CRAH unit supply air temperature setpoints, and their fan speed ratios. For benchmarking, we use a conventional rule-based strategy that adheres to widely accepted industry standards for cooling operation. According to SLA constraints, the inlet temperature of the IT equipment must be kept between 18 °C and 27 °C, while the relative humidity should remain within 30\% to 60\%. Any breach of these constraints during training or evaluation incurs penalties to ensure safety-compliant optimization behavior. To model the thermal behavior of the system with high fidelity, we adopt the widely used simulation engine EnergyPlus~\cite{crawley2001energyplus}, which enables detailed representation of heat transfer dynamics and equipment performance. Additionally, the Tianshou library~\cite{tianshou} is used to construct the reinforcement learning policy reservoir.

\subsection{Digital Twin Construction}

\begin{figure}[!t]
    \centering
    \includegraphics[width=0.7\textwidth]{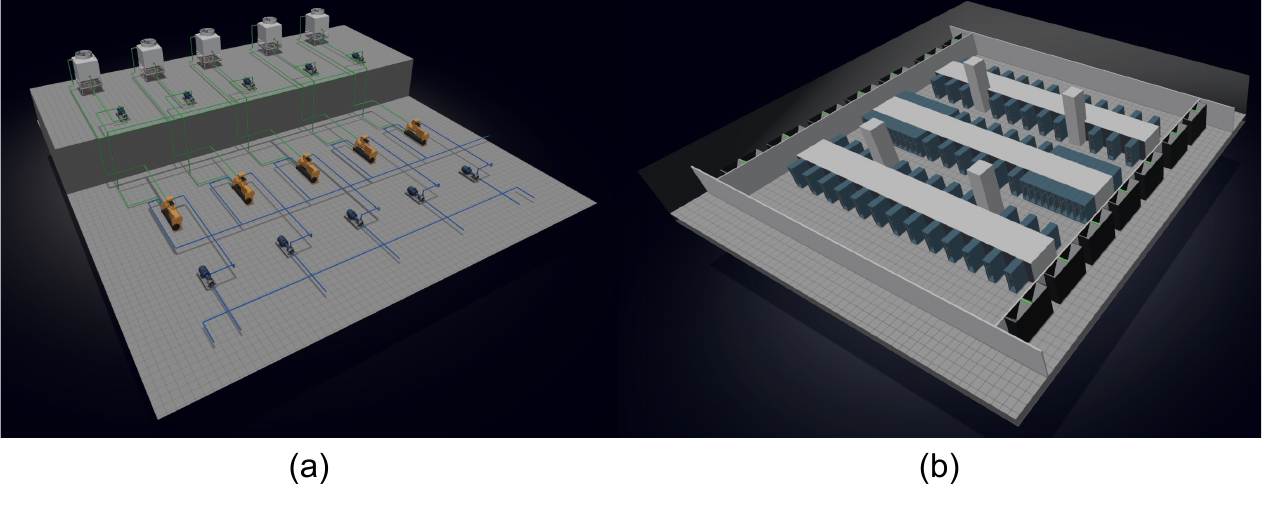}
    \caption{Digital twin layout. (a) Cooling system. (b) Data hall.}
    \label{DTLayout}
\end{figure}

The digital twin of the data center’s data hall and cooling infrastructure was developed through a systematic workflow integrating geometric reconstruction, physics-based thermodynamic modeling, data-driven surrogate modeling, and calibration using field data. Based on the actual facility layout (Fig.~\ref{DataCenterLayout}), a geometry-informed digital replica of the data hall and cooling system was constructed, as illustrated in Fig.~\ref{DTLayout}. This representation encodes key spatial attributes such as rack placement, airflow organization under cold aisle containment, and the positioning of major cooling components, including CRAH units, CHW loops, and cooling towers.

Subsequently, physics-informed models were established for individual cooling system components based on manufacturer specifications. These models characterize the operational behavior and design constraints of equipment such as cooling towers, condenser water pumps, CHW pumps, chillers, and CRAH fans. Their interactions were simulated to yield a high-resolution depiction of system dynamics under various control inputs and environmental conditions. To facilitate efficient calibration, a data-driven surrogate modeling strategy was employed. Key control variables—including CHW supply temperature, CRAH supply air temperature setpoints, and fan speed ratios—were systematically perturbed, and the corresponding system responses were obtained via EnergyPlus simulations. A lightweight surrogate model was then trained on the resulting dataset to approximate the input-output mappings with significantly reduced computational cost.
\begin{figure*}[!t]
    \centering
    \includegraphics[width=\textwidth]{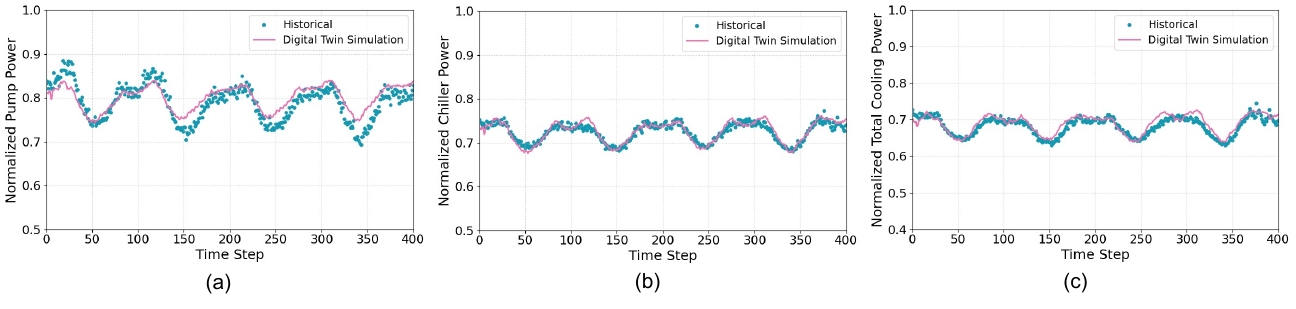}
    \caption{Normalized digital twin cooling system calibration results. (a) Total chilled water pump power. (b) Total chiller power. (c) Total cooling plant power.}
    \label{CalibrationResults}
\end{figure*}
\begin{table}[!t]
\centering
\caption{Model prediction accuracy for key features.}
\label{tab:feature_mape}
\begin{tabularx}{\linewidth}{X >{\raggedright\arraybackslash}X}
\toprule
\textbf{Feature} & \textbf{MAPE} \\
\midrule
CRAH averaged return air temperature & 0.97\% \\
Chilled water return temperature & 0.50\% \\
Chiller plant total cooling load & 1.58\% \\
Total chilled water pump power & 2.89\% \\
Total chiller power & 1.11\% \\
Total cooling plant power & 1.86\% \\
\bottomrule
\end{tabularx}
\end{table}

Finally, the historical dataset was used to perform the calibration of the digital twin. The calibrated digital twin was aligned with real-world measurements, minimizing prediction errors across key observables. To quantitatively evaluate the calibration accuracy, we employed the Mean Absolute Percentage Error (MAPE) metric, defined as:
\begin{equation}
    \text{MAPE} = \frac{1}{N} \sum_{i=1}^{N} \left| \frac{y_i - \hat{y}_i}{y_i} \right| \times 100\%,
\end{equation}
where \( y_i \) denotes the observed value, \( \hat{y}_i \) denotes the predicted value from the digital twin, and \( N \) is the total number of evaluation samples. The calibration results are illustrated in Fig.~\ref{CalibrationResults}. The MAPE for the key features is summarized in Table~\ref{tab:feature_mape}. The results show that the calibrated digital twin achieves high accuracy in predicting both thermal and energy dynamics, thereby ensuring its reliability as a virtual environment for DRL training, validation, and deployment support.

\subsection{Deep Reinforcement Learning Pre-Evaluation}
Leveraging the calibrated digital twin as a computationally efficient training environment, we conducted extensive training of multiple DRL agents. The hybrid modeling structure of the digital twin enabled low-overhead exploration across different control strategies, DRL algorithms, and hyperparameter configurations. Specifically, we compared policies optimizing only CRAH units (supply air temperature and fan speed) with those jointly optimizing CRAH units and CHW supply temperature and systematically evaluated multiple DRL algorithms under diverse hyperparameter settings. Subsequently, an initial policy reservoir comprising various candidate policies was constructed. To ensure safe and effective deployment, we filtered this reservoir based on predefined criteria, specifically energy efficiency improvements and adherence to SLA constraints. Candidate policies demonstrating poor convergence, excessive energy consumption, or violations of operational constraints were excluded.

\begin{figure*}[!t]
    \centering
    \includegraphics[width=\textwidth]{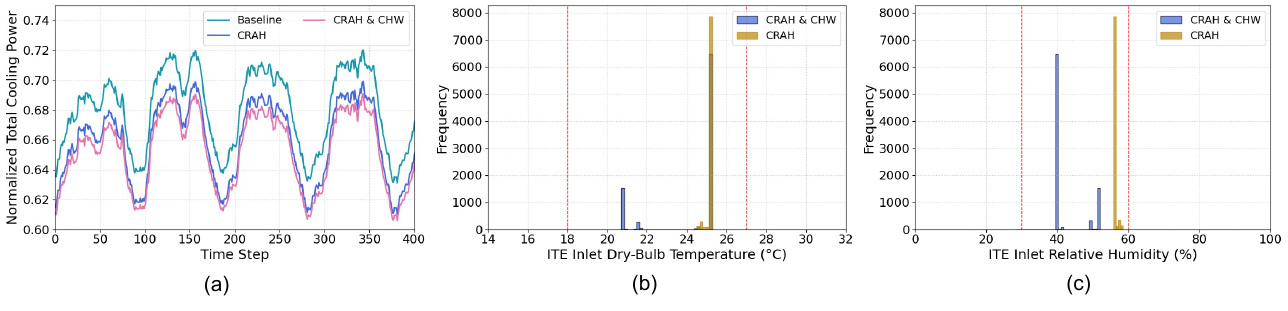}
    \caption{Comparisons of various control strategies. (a) Normalized total power consumption. (b) SLA: ITE inlet dry-bulb temperature. (c) SLA: ITE inlet relative humidity.}
    \label{AIPowerSLA}
\end{figure*}

The filtered candidate policies were then evaluated with the digital twin to identify the optimal control strategy. The performance comparisons, depicted in Fig.~\ref{AIPowerSLA}, illustrate notable power savings relative to the baseline rule-based strategy. Specifically, the CRAH-only optimization strategy “CRAH” achieved approximately 3.08\% energy savings. The joint optimization “CRAH \& CHW” further enhanced performance, achieving approximately 4.09\% energy savings compared to the “Baseline”. Importantly, both optimized strategies consistently satisfied the stringent SLA constraints, maintaining the IT equipment inlet air temperature within 18°C–27°C and relative humidity within 30\%–60\% throughout the evaluation period, demonstrating full compliance (100\%).
\begin{figure*}[!t]
    \centering
    \includegraphics[width=\textwidth]{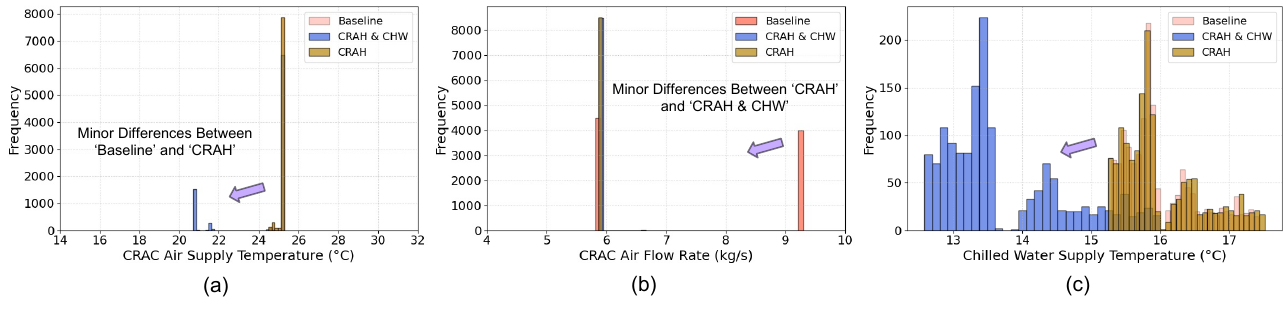}
    \caption{Control actions of various control strategies. (a) CRAH air supply temperature. (b) CRAH air flow rate. (c) Chilled water supply temperature.}
    \label{ControlActionCmparison}
\end{figure*}

Leveraging the digital twin model, we further analyzed DRL agent behaviors to enhance the interpretability of AI-based control policies. As an illustrative example, Fig.~\ref{ControlActionCmparison} presents histograms comparing control action distributions under three strategies: “Baseline”, “CRAH”, and “CRAH \& CHW”. From Fig.~\ref{ControlActionCmparison}, the three strategies exhibit similar distributions for CRAH supply air temperature setpoints, although the joint “CRAH \& CHW” strategy occasionally selects lower setpoints (20–22°C) under specific conditions. For CRAH fan speed ratios, both AI-based strategies favor lower airflow rates than the baseline while maintaining SLA compliance. Regarding chilled water supply temperature, the “Baseline” and “CRAH” strategies show minimal differences, as neither actively controls this parameter. In contrast, the joint optimization strategy consistently selects lower chilled water supply temperatures. These results illustrate differences in the control actions across strategies, though the precise implications of these actions on individual cooling system components require further detailed investigation.

\begin{figure*}[!t]
    \centering
    \includegraphics[width=\textwidth]{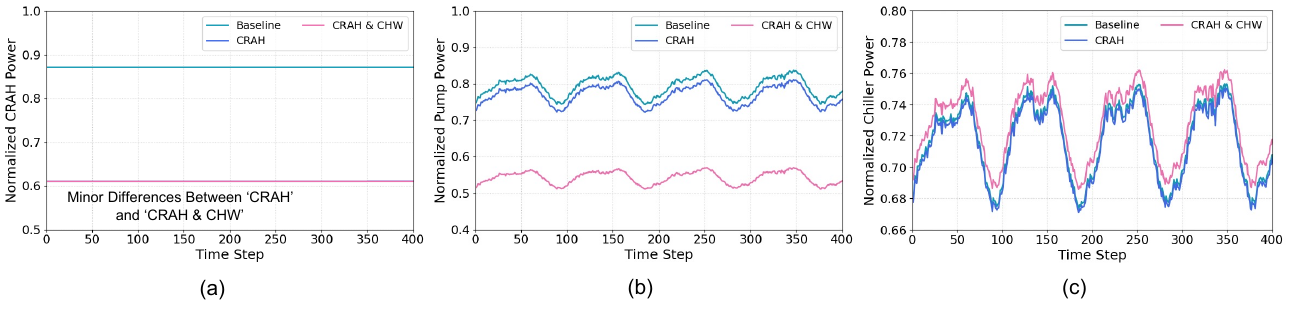}
    \caption{Normalized power consumption of cooling system equipment. (a) CRAHs. (b) CHW pumps. (c) Chillers.}
    \label{PowerBreakdown}
\end{figure*}

Fig.~\ref{PowerBreakdown} compares the power consumption of key cooling system components under three control strategies. As shown in Fig.~\ref{PowerBreakdown}(d) and (e), the condenser water pump and cooling tower exhibit minimal power consumption differences across strategies since they were not directly optimized. Fig.~\ref{PowerBreakdown}(a) demonstrates a significant reduction in CRAH fan power consumption under both the ``CRAH'' and ``CRAH\&CHW'' strategies compared to the ``Baseline''. Combined with the action analysis in Fig.~\ref{ControlActionCmparison}(b), this energy saving primarily results from the AI strategies' preference for lower supply airflow rates. Moreover, as depicted in Fig.~\ref{ControlActionCmparison}(c), the ``CRAH\&CHW'' strategy selects a lower chilled water supply temperature, leading to reduced chilled water flow for the same cooling load and thus lower chilled water pump energy consumption (Fig.~\ref{PowerBreakdown}(b)). Although the lower chilled water temperature increases the chiller power usage (Fig.~\ref{PowerBreakdown}(c)), the overall system energy consumption decreases. This explains why the ``CRAH\&CHW'' strategy achieves superior energy savings compared to the ``CRAH'' strategy as indicated in Fig.~\ref{AIPowerSLA}(a).

In summary, the digital twin platform provides an efficient and safe playground for training a wide range of DRL policies with diverse control actions, algorithmic choices, and hyperparameter settings. It enables the construction of a high-quality policy reservoir and supports rigorous pre-evaluation of candidate policies before real-world deployment. Additionally, the analysis of control behaviors within the digital twin enhances the interpretability of AI decision-making, thereby increasing operator trust of AI-driven data center control systems.

\subsection{Limitation and Future Works}

While the proposed DLCF demonstrates significant advantages in improving the reliability of DRL deployment in data centers, several limitations remain, pointing toward promising future research directions. First, from a theoretical perspective, the deployment of DRL algorithms within the dual-loop framework currently lacks a comprehensive and rigorous theoretical foundation. Although this work analyzes key aspects such as sample complexity, generalization, optimality, and safety, a more complete and unified theory that formally characterizes the interactions between DRL agents, digital twins, and physical systems remains an open research challenge. Establishing such a theory would provide stronger guarantees for performance and safety during real-world deployment. Second, building on the theoretical foundations, future work should focus on further advancing DRL algorithms tailored to the dual-loop setting. New algorithmic designs that explicitly account for digital twin model uncertainties and enable online adaptation to real-time data assimilation could further strengthen the reliability and effectiveness of DRL control in complex cyber-physical environments. Third, the current study validates the proposed framework using a high-fidelity digital twin based on a real-world data center setup. However, broader experimental validation across diverse, large-scale real-world data centers remains necessary to assess the reliability. Fourth, this study primarily focuses on optimizing the cooling subsystem. Scaling the framework to optimize multiple interconnected subsystems—such as workload management, energy storage, and renewable energy integration—poses greater challenges in terms of coordination, cross-domain interaction modeling, and system complexity. Future work could explore hierarchical RL, multi-agent RL, and system-wide optimization strategies to address the higher dimensionality and coupled dynamics of full-stack data center control.

\section{Conclusion}
\label{sec:conclusion}
This paper presents the DLCF, a digital twin–enabled architecture designed to facilitate the reliable deployment of DRL in data centers. By tightly integrating physical systems, digital twins, and DRL policy reservoirs through two interconnected loops, DLCF aims to address key limitations in existing DRL deployment paradigms, including data scarcity and inadequate real-time evaluation mechanisms.

We conducted theoretical analyses to demonstrate how the proposed design improves sample efficiency, generalization, optimality, and safety. Case studies on the digital twin of a real-world data center cooling system validate the practical effectiveness of the framework. The results demonstrate up to 4.09\% energy savings compared to conventional control strategies, with full compliance to SLA constraints. More importantly, it demonstrates how the DLCF enhances the interpretability and trustworthiness of AI-driven control.

Regarding further works, a unified theoretical foundation is needed to rigorously characterize the interplay between DRL agents, digital twins, and physical systems. Furthermore, extending DLCF beyond cooling control to support system-wide optimization—encompassing workload management, energy storage, and renewable integration—offers a promising direction toward building intelligent and sustainable next-generation data centers.


\bibliographystyle{unsrt}
\bibliography{cas-refs}
\end{document}